\documentclass[11pt]{article}

\usepackage[preprint]{acl}

\usepackage{times}
\usepackage{latexsym}

\usepackage[T1]{fontenc}

\usepackage[utf8]{inputenc}

\usepackage{microtype}

\usepackage{inconsolata}

\usepackage{graphicx}

\usepackage{amsmath}

\usepackage{xcolor}

\usepackage{booktabs}

\usepackage{amsfonts} 

\usepackage[dvipsnames]{xcolor}

\usepackage{subcaption} 

\usepackage{array}

\usepackage{tcolorbox}
\tcbuselibrary{skins}

\usepackage{tikz}
\usetikzlibrary{arrows.meta, positioning}

\usepackage{CJKutf8}

\title{Seeing Isn’t Resolving: \\ Visual and Linguistic Disambiguation with Conversational Grounding}

\author{
    \textbf{Edoardo Cecchinato\textsuperscript{1}} \quad
    \textbf{Shun Inadumi\textsuperscript{2}} \quad
    \textbf{Angel Garc\'ia Contreras\textsuperscript{3}}
    \\
    \textbf{Giuseppe Riccardi\textsuperscript{1}} \quad
    \textbf{Koichiro Yoshino\textsuperscript{2,3}}
    \\[0.2em]
    \textnormal{\textsuperscript{1}Signals and Interactive Systems Lab, University of Trento, Italy}
    \\
    \textnormal{\textsuperscript{2}The University of Osaka, Japan}
    \textnormal{\textsuperscript{3}Guardian Robot Project, RIKEN, Japan}
    \\
    {\texttt{edoardo.cecchinato@unitn.it}}
}

\begin{document}
\maketitle
\begin{abstract}
Visual grounding benchmarks usually do not consider the contextual uncertainty of real-world scenarios. In this study, we construct a novel benchmark, the Japanese evaluation for Ambiguous visual Grounding with Dialogue (JAG-D) dataset. It consists of human-robot conversational multimodal interactions covering different types of ambiguity, arising from visual and linguistic categories. We evaluate the capabilities of different state-of-the-art Visual Language Models (VLMs) for ambiguity recognition and resolution. We compare dialogue context grounding performance for visual and linguistic ambiguity resolution. Our experiments show that VLMs can substantially improve the resolution of linguistic ambiguity, especially Japanese zero anaphora, by using dialogue context. We further conduct a comparative analysis to understand whether grounding improvements arise from the semantic content of the dialogue or from its conversational structure\footnote{We release \href{https://github.com/riken-grp/seeing_is_not_resolving}{code} and \href{https://huggingface.co/datasets/edo00/JAG-D}{data}.}.
\end{abstract}

\section{Introduction}

\begin{figure}[t!]
    \centering
    \includegraphics[width=\linewidth]{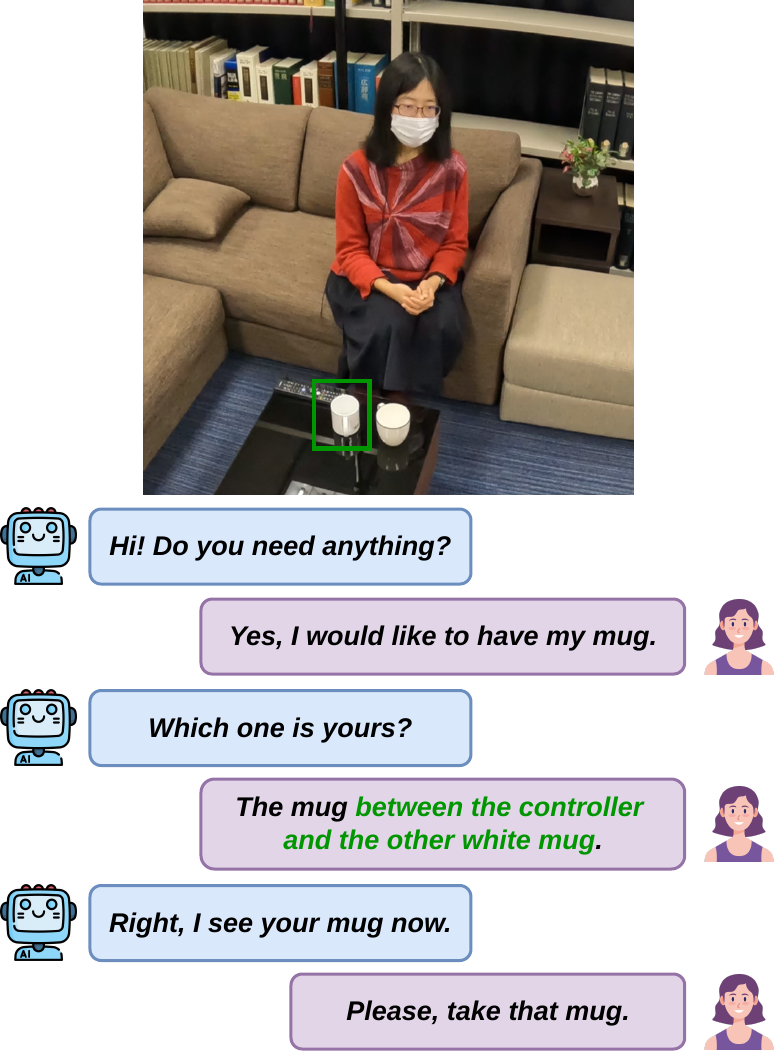}
    \caption{Example of visual and lingusitic ambiguity resolution through dialogue history integration, where dialogue turns provide additional context to the ambiguous deictic instruction (\textit{take that mug}). The dialogue text has been translated from Japanese, the original language of the human-robot interactions.}
    \label{fig:first_page}
\end{figure}

Visual grounding is the task of localizing visual regions described by referring expressions in natural language, and has achieved substantial progress on standard benchmarks \citep{Plummer_2015_ICCV, Rohrbach_2016}. However, these benchmarks evaluate models under idealized conditions where referring expressions are unambiguous, single-target, and context-independent \citep{11235566}, even if such assumptions are not valid for real-world settings. This limitation motivates the need to address ambiguity in referring expressions in realistic scenarios such as human-robot interaction.

Ambiguity resolution in language has been widely studied, particularly at the lexical and sentential levels \citep{gonzalo-etal-1999-lexical, liu-etal-2023-afraid, schouten-etal-2023-reasoning, karamolegkou-etal-2025-trick}. Multimodal ambiguity resolution is equally important, and tasks such as visual question answering have attracted particular attention \citep{stengel-eskin-etal-2023-chicken, Nam_2025_ICCV, zhou-etal-2025-focus}. However, such tasks capture only part of the referential communication required in human-robot interaction. To achieve appropriate interaction grounded to the world, computational systems must be able to properly interpret references that depend more heavily on the situational context. In real-world scenarios, user utterances contain diverse forms of instructions and omissions, and resolving them requires the use of various types of information, such as dialogue history~\cite{devault2009learning}, pointing~\cite{constantin2022interactive}, and gaze~\cite{inadumi-etal-2024-gaze}. In particular, appropriately interpreting context by combining dialogue with visual input is one of the capabilities required of dialogue agents \citep{Das_2017_CVPR}.

For example, consider the situation shown in Figure \ref{fig:first_page}, where two mugs are placed side by side on a table in a living room and a human instructs a robot to ``take that mug.’’ To resolve this ambiguity, it is necessary to use the dialogue context that indicates what ``that’’ refers to. In this case, the ambiguity to be resolved through the dialogue context exists in both the visual and linguistic modalities. Visual ambiguity can arise when multiple similar objects are present in the visual scene, or when multiple objects share the same attributes \citep{Zhang_2018_CVPR, Zeng-etal-2024_CVPR}. Linguistic ambiguity \citep{berzak-etal-2015-see}, on the other hand, prominently involves demonstratives, overt pronouns, and zero pronouns (zero anaphora). Zero anaphora is a linguistic phenomenon in which a referring expression, such as a subject or object, is omitted but recoverable from context. It is particularly prevalent in pro-drop languages such as Japanese~\cite{seki-etal-2002-probabilistic}, Chinese~\cite{kong-zhou-2010-tree}, and Korean~\cite{park-etal-2015-zero}. In this work, we especially focus on phenomena in Japanese.

Existing benchmarks do not jointly control the source of ambiguity and the context used to resolve it, making it impossible to determine whether grounding failures arise from perceptual or discourse-level limitations. To clarify the extent to which multimodal systems can address these problems, and where their limitations lie, we introduce Japanese evaluation for Ambiguous visual Grounding with Dialogue (JAG-D). JAG-D contains 480 carefully curated instances, each combining an image, an instruction, and a dialogue history. We prioritize annotation quality over benchmark scale, limiting the dataset size so that human experts, with more than eight years of experience in discourse structure annotation, can thoroughly verify each instance. The utterances in JAG-D contain diverse ambiguities arising from both visual and linguistic factors, and the dataset consists of situations in which these ambiguities are expected to be resolved by referring to the dialogue history.

Using this benchmark, we discuss the ambiguity recognition and ambiguity resolution capabilities of visual grounding systems, including current VLMs. First, we evaluate the ability to determine whether a given pair of current instruction and image situation contains ambiguity. Second, we discuss ambiguity resolution ability when dialogue context is provided, namely the improvement in grounding accuracy. As evaluation targets, we use representative visual grounding models including GLIP~\cite{Li2021GroundedLP} and MDETR~\cite{Kamath2021MDETRM}, as well as Qwen3-VL~\cite{bai2025qwen3vltechnicalreport}, a general-purpose VLM with visual grounding capabilities.

Our experiments reveal that current models can robustly recognize linguistically ambiguous scenarios; however, fail when the ambiguity arises from the visual scene. We also find that using additional dialogue context substantially resolves linguistic ambiguity, especially ambiguity caused by zero anaphora, whereas its contribution to resolving visual ambiguity is limited. This asymmetry suggests that current multimodal models rely heavily on linguistic surface cues rather than actively reasoning about the structure of the visual scene. Closing this gap will require models that treat perception and discourse as jointly informative sources \citep{inadumi-etal-2025-disambiguating}, rather than treating dialogue as a post-hoc filter over linguistically driven predictions.

Our contributions are as follows:
\begin{enumerate}
\setlength{\parskip}{0cm}
\setlength{\itemsep}{0cm}

\item We introduce JAG-D, a Japanese benchmark for studying ambiguity resolution in human-robot visual grounding. JAG-D provides controlled coverage of diverse linguistic and visual ambiguities, as well as dialogue history used as disambiguation context.
\item We provide a systematic evaluation of ambiguity recognition and ambiguous visual grounding. Our results show that visual ambiguity remains a systematic blind spot for both monolingual Japanese and multilingual VLMs.
\item We show that dialogue history improves visual grounding by resolving ambiguities that cannot be addressed from a single instruction alone. We further introduce a dialogue intervention analysis to quantify how different dialogue aspects affect grounding accuracy.
\end{enumerate}

\section{Related Work}
\paragraph{Visual Grounding.}
Visual grounding is a fundamental task that links referring expressions to regions in the visual world. The literature includes foundational benchmarks like RefCOCO and its variations \citep{7780378, kazemzadeh-etal-2014-referitgame, 10.1007/978-3-319-46475-6_5}, Flickr30k Entities \citep{Plummer_2015_ICCV}, and Visual Genome \citep{krishna2017visual}, alongside an evolution of model development: from early methods based on alignment \citep{karpathy2015deep, Rohrbach_2016} to transformer architectures like MDETR \citep{Kamath2021MDETRM}, GLIP \citep{Li2021GroundedLP}, and GroundingDINO \citep{Liu2025GroundingDINO} that achieve strong performance on the grounding task. Additionally, several studies have also explored extensions of visual grounding to more complex settings, such as including pronouns in addition to noun phrases \citep{lu-etal-2022-extending} and grounding in contexts of human-robot interaction \citep{Mi2020GroundingRobotics}. Despite these advances, most existing benchmarks still lack a well-defined categorization of ambiguity and real-world conversations for human-robot interaction.

\paragraph{Ambiguity in Vision–Language Tasks.}
Ambiguity is widely recognized as a fundamental challenge in multimodal understanding. Prior work has primarily studied this phenomenon in tasks such as visual question answering (VQA) \citep{inadumi-etal-2024-gaze, testoni-etal-2025-racquet, jian-etal-2025-teaching, jang2026aqua}, as well as multimodal machine translation \citep{ma-etal-2024-3am}. In human-robot interaction, systems are required not only to answer questions but also to localize and use intended objects in the visual scene. While these studies partially address such challenges, they do not sufficiently explore ambiguity that can arise from visual similarity or linguistic underspecification in dialogue contexts.

\paragraph{Visual Grounding with Context.}
Integrating dialogue as additional context for visual grounding has also been explored. Benchmarks like SIMMC 2.0 \citep{kottur-etal-2021-simmc}, \textit{GuessWhat?!} \citep{Tu_2021_CVPR} and \textit{InViG} \citep{Zhang_2024_CVPR} contain dialogues with disambiguation questions, allowing model evaluation under open-ended ambiguous queries. As a challenging setting, the J-CRe3 dataset \citep{ueda-etal-2024-j} provides visual grounding in Japanese dialogue as one of its tasks, offering insight into visual grounding under multimodal ambiguity. Our benchmark extends J-CRe3 and investigates more practical ambiguities that depend on the context of human-robot interaction.

\section{Problem Definition}

Let $I$ be a human-robot interaction image from robot's first-person view, and $u$ a textual human-to-robot instruction, potentially ambiguous due to multiple possible references in the image. Additionally, let $H = \{u_1,\dots,u_K\}$ be a $K$-turns dialogue history that contains the human-robot exchanges $u_i$ leading to the instruction $u$. Finally, $b^* \in \mathcal{B}(I)$ is a single ground-truth bounding box that contains the target object for the grounding task, where $\mathcal{B}(I)$ denotes the set of candidate bounding boxes in $I$.

\subsection{Task Formulation}
\label{sec:task_formulation}

\paragraph{Multimodal Ambiguity Recognition.}
We evaluate models on ambiguity recognition, defined as predicting whether $(I,u)$ is ambiguous, i.e., whether the instruction $u$ alone is not sufficient to correctly identify a clear target object $\hat{b}$ in image $I$. The models receive as input the pair $(I,u)$ and output \textit{``ambiguous''} or \textit{``non-ambiguous''} based on their processing.

Let $\phi(u,I,b) \in \{0,1\}$ denote whether region $b$ in image $I$
satisfies the semantic interpretation of instruction $u$.
We define
\[
\mathcal{R}(u,I) = \{ b \in \mathcal{B}(I) \mid \phi(u,I,b) = 1 \}
\]

as the set containing all the bounding boxes that ground a reference in the instruction $u$.

Although supervision provides a single intended referent $b^*$, ambiguity is defined independently of annotation, and for an image-instruction input pair $(u,I)$ the possible outcomes arise:

\[
y =
\begin{cases}
1 & \text{if } |\mathcal{R}(u,I)| > 1 \quad (\text{ambiguous}) \\
0 & \text{if } |\mathcal{R}(u,I)| = 1 \quad (\text{unambiguous})
\end{cases}
\]

We model ambiguity recognition as a binary classification task.
Given $(I,u)$, a classifier
\[
c(I,u) \in [0,1]
\]
predicts the probability that the instruction is ambiguous:
\[
\hat{y} = \mathbb{I}[c(I,u) > \tau],
\]
where $\tau$ is a decision threshold.

\paragraph{Visual Grounding Evaluation.}
In addition to the recognition task, we study visual grounding under ambiguous input in real-world conversations. Given an image–instruction pair $(I,u)$, the grounding model predicts a bounding box $\hat{b}$ as follows:
\[
\hat{b} = f(I, u), \quad \hat{b} \in \mathcal{B}(I)
\]
The performance is evaluated using Intersection over Union (IoU) between $\hat{b}$ and the ground-truth bounding box $b^*$ \citep{everingham2010pascal}.

For the visual grounding evaluation, we consider two settings:
\begin{enumerate}
\setlength{\parskip}{0cm}
\setlength{\itemsep}{0cm}

    \item \textbf{Instruction-only grounding:} The model receives the instruction $u$ only.
    \item \textbf{Dialogue-enhanced grounding:} The dialogue history $H$ is integrated with the instruction $u$ to obtain a disambiguated instruction $\hat{u}$ as follows:
\[
\hat{u} = g(u, H),
\]
where $g$ is the dialogue-integration function that processes the information from $H$ and uses it to better interpret $u$.
\end{enumerate}

\subsection{Ambiguity Categorization}
\label{sec:ambiguity_taxonomy}

In our work, we assume two types of ambiguity that can occur in multimodal inputs to conversational systems. These ambiguity categories are used to balance the data during dataset construction and are also used in the subsequent analysis.

\paragraph{Visual Ambiguity.}
The instruction $u$ is linguistically explicit, but multiple objects in image $I$ match the description. We distinguish two types of visual ambiguity:
\begin{itemize}
\setlength{\parskip}{0cm}
\setlength{\itemsep}{0cm}
    
    \item \textbf{Same Object:} multiple instances of the same object candidate, for instance, \textit{``pass me the \textcolor{BrickRed}{mug}''} (\textit{``\begin{CJK}{UTF8}{ipxm}\textcolor{BrickRed}{マグカップ}を取って\end{CJK}''}) where there is more than one mug in the image.
    
    \item \textbf{Same Attribute:} multiple objects share the specified attribute(s), such as \textit{``pass me the \textcolor{BrickRed}{white} object''} (\textit{``\begin{CJK}{UTF8}{ipxm}\textcolor{BrickRed}{白い}物体を取って\end{CJK}''}) where there is more than one white candidate in the visual input.
    
\end{itemize}

\paragraph{Linguistic Ambiguity.}
The instruction $u$ is underspecified at the discourse level, as it lacks sufficient contextual information to identify the target object in the image $I$ uniquely. Consequently, we categorize it as follows:
\begin{itemize}
\setlength{\parskip}{0cm}
\setlength{\itemsep}{0cm}
    
    \item \textbf{Pronoun Ambiguity:} the use of an explicit pronoun that lacks a clear antecedent within the current utterance, such as \textit{``put \textcolor{BrickRed}{it} on the table''} (\textit{``\begin{CJK}{UTF8}{ipxm}テーブルに\textcolor{BrickRed}{それ}を置いて\end{CJK}''}) where \textit{``it''} refers to a previously discussed object.
    
    \item \textbf{Zero Anaphora:} an omitted argument that requires contextual recovery, such as \textit{``\textcolor{BrickRed}{pass}''} (\textit{``\textcolor{BrickRed}{\begin{CJK}{UTF8}{ipxm}取って\end{CJK}}''}). Zero anaphora is very common in Japanese language and requires the dialogue history $H$ to be resolved.
    
\end{itemize}

\section{JAG-D Overview}
\label{sec:data_overview}

\begin{table}[t]
\centering
\small
\begin{tabular}{@{}lcc@{}}
\toprule
\textbf{Category} & \textbf{Count} & \textbf{\%} \\
\midrule
\multicolumn{3}{@{}l}{\textbf{Instruction Ambiguity}} \\
Ambiguous & 232 & 48.3 \\
\quad \textit{same\_object} & 58 & 12.1 \\
\quad \textit{same\_attribute} & 55 & 11.5 \\
\quad \textit{pronoun} & 61 & 12.7 \\
\quad \textit{zero\_anaphora} & 58 & 12.1 \\
Unambiguous & 248 & 51.7 \\
\midrule
\multicolumn{3}{@{}l}{\textbf{Dialogue Mode}} \\
Resolving & 174 & 36.2 \\
Non-resolving & 118 & 24.6 \\
Supportive & 188 & 39.2 \\
\bottomrule
\end{tabular}
\caption{Distribution of the JAG-D benchmark ($N=480$). Instruction ambiguity is divided into unambiguous and ambiguous types, with the latter spanning four categories. Dialogue modes are assigned as follows: \textit{resolving} for ambiguous instructions, \textit{supportive} for unambiguous instructions, and \textit{non-resolving} for both.}
\label{tab:dataset_distribution}
\end{table}

We present JAG-D, a Japanese benchmark for evaluating visual grounding under controlled multimodal ambiguity in conversational settings. The dataset isolates ambiguity categories and explicitly measures both ambiguity recognition and dialogue-based resolution. We focus on Japanese because zero anaphora is frequently used in the language, making it natural for linguistic ambiguity to arise in ways that strongly depend on dialogue context \citep{ono1997deconstructing, seki-etal-2002-probabilistic}.

\subsection{Source Images}

The images in the test set are sampled from J-CRe3 \citep{ueda-etal-2024-j}, an egocentric human-robot interaction dataset collected in real-world environments (e.g., kitchens, offices, living rooms). We construct 480 instances starting from J-CRe3 visual content, prioritizing scenes with sufficient object density and diversity to enable controlled ambiguity under different environment combinations.

\subsection{Design Principles}

\paragraph{Controlled Multimodal Ambiguity.}
Each instance contains a single ground-truth bounding box $b^*$. Ambiguity is introduced by ensuring that multiple regions satisfy the semantic constraints of instruction $u$, while only one corresponds to the intended referent.

\paragraph{Explicit Ambiguity Categorization.}
Ambiguous instructions are categorized into four types, described in Section~\ref{sec:ambiguity_taxonomy}: \textit{same\_object}, \textit{same\_attribute}, \textit{pronoun\_ambiguity}, and \textit{zero\_anaphora}. This categorization enables fine-grained evaluation across distinct variants of ambiguity.

\paragraph{Dialogue-Guided Disambiguation.}
Each instance consists of an instruction $u$ and dialogue history $H$ under three modes:
(i) \emph{resolving}, where $H$ disambiguates $u$,
(ii) \emph{non-resolving}, where ambiguity persists due to irrelevant context,
(iii) \emph{supportive}, where $H$ provides redundant but consistent information for an unambiguous $u$.
Examples are provided in Appendix \ref{sec:appendix_dialogue_mode}.

\subsection{Data Generation and Validation}

For each image, we generate an ambiguous or unambiguous instruction paired with dialogue history consistent with the designated ambiguity type and dialogue mode. The data generation pipeline is described in Appendix~\ref{sec:appendix_data_generation}.

All instances are manually reviewed and corrected by two native Japanese annotators, with more than eight years of experience, ensuring that (i) the ambiguity matches the assigned category, (ii) the dialogue mode functions as intended, and (iii) the target bounding box aligns with the instruction. 

The final benchmark comprises 480 instances with balanced ambiguous and unambiguous conditions and controlled distributions across ambiguity types and dialogue settings (Table~\ref{tab:dataset_distribution}). JAG-D provides a carefully curated, human-evaluated test set of high-quality real-world disambiguation dialogues.

\section{Experimental Setup}

\subsection{Multimodal Ambiguity Recognition}

We evaluate Japanese and multilingual VLMs on the multimodal ambiguity recognition task, using our benchmark. Specifically, we provide the models with two inputs: an image and an instruction, where the instruction is either ambiguous or unambiguous according to our ambiguity categorization.

The models tested are Japanese VLMs, i.e., LLM-jp VILA\footnote{\url{https://huggingface.co/llm-jp/llm-jp-3-vila-14b}} \citep{sasagawa-etal-2025-constructing}, Sarashina2\footnote{\url{https://huggingface.co/sbintuitions/sarashina2-vision-8b}}, and EvoVLM-JP\footnote{\url{https://huggingface.co/SakanaAI/Llama-3-EvoVLM-JP-v2}}~\cite{Akiba_2025}, and two multilingual VLMs, i.e., Qwen3-VL\footnote{\url{https://huggingface.co/Qwen/Qwen3-VL-8B-Instruct}}~\cite{bai2025qwen3vltechnicalreport} and Gemma3\footnote{\url{https://huggingface.co/google/gemma-3-12b-it}}~\cite{gemma2025}. They were prompted under different settings (zero-shot and few-shot) and in both Japanese and English for the multilingual models (the prompts are provided in Appendix~\ref{sec:appendix_amb_rec}).

We evaluate the models on macro F1 score for the overall recognition performance, and we also compute per-category recall following our benchmark's ambiguity categorization.

\subsection{Visual Grounding Evaluation}

We evaluate models on ambiguous visual grounding settings with dialogue history integration as disambiguation context. Our evaluation includes the Japanese versions of two widely used grounding models, GLIP and MDETR, trained on both English datasets~\cite{krishna2017visual, Hudson_2019_CVPR} and Japanese datasets \citep{nakayama-etal-2020-visually, ueda-etal-2024-j}. We also evaluate Qwen3-VL, which has demonstrated strong multilingual performance and visual grounding capabilities \citep{bai2025qwen3vltechnicalreport}.

The evaluation pipeline comprises a two-phase evaluation protocol, designed to isolate grounding performance from raw user instruction (first phase) and the benefit of additional dialogue context to solve ambiguous scenarios (second phase). From the ones mentioned, Qwen is the only model that can handle long input, such as a complete dialogue history, while standard visual grounding models usually process short sentences or a list of phrases only. Consequently, we leverage an established method in the literature, by \citet{willemsen-etal-2023-resolving}, to integrate the dialogue information into a single simple input sentence for the second-phase grounding task for GLIP and MDETR. This integration method consists of using an LLM (GPT-5.2\footnote{\href{https://developers.openai.com/api/docs/models/gpt-5.2}{gpt-5.2-2025-12-11}} in our case) to retrieve the important information from a dialogue and condense it into a shorter textual input. Both the Qwen's prompt and the one used for the dialogue integration can be found in Appendix~\ref{sec:appendix_grounding}.

The metrics used in the visual grounding evaluation are mean Intersection over Union (mIoU) on the top-1 prediction and Acc@$\tau$, measuring if the model's top-scoring bounding box attains IoU $\geq \tau$, with $\tau$ of 0.5 and 0.75 in our experiments.

\begin{table*}[t]
\centering
\small
\begin{tabular}{lcc|cc|cc}
\toprule
& & & \multicolumn{2}{c|}{\textbf{Visual Ambiguity (R)}} & \multicolumn{2}{c}{\textbf{Linguistic Ambiguity (R)}} \\
\textbf{Model} & \textbf{Overall (F1-M)} & \textbf{Non-Ambig (R)} & \textbf{Same Obj} & \textbf{Same Attr} & \textbf{Pronoun} & \textbf{Zero Ana} \\
\midrule
Qwen3-VL & \textbf{83.0} & 94.4 & 32.8 & 54.5 & 96.7 & 100.0 \\
Gemma3 & 78.5 & 85.9 & 39.7 & 52.7 & 96.7 & 93.1 \\
LLM-jp VILA & 60.1 & 67.3 & 46.6 & 54.5 & 47.5 & 63.8 \\
Sarashina2 & 40.3 & \textbf{\textcolor{BrickRed}{7.7}} & 98.3 & 98.2 & 100.0 & 100.0 \\
EvoVLM-JP & 32.6 & \textbf{\textcolor{BrickRed}{0.0}} & 100.0 & 100.0 & 100.0 & 100.0 \\
\bottomrule
\end{tabular}
\caption{Multimodal ambiguity recognition on the full test set ($N{=}480$; binary ambiguous vs.\ non-ambiguous). We report macro-averaged F1 (\textbf{F1-M}) for the overall results and recall (\textbf{R}) for the category-specific performance. \textbf{Bold} represents higher overall performance. The \textbf{\textcolor{BrickRed}{red color}} represents models that are highly biased toward the ambiguous label. This table reports the few-shot models' results only, while a complete results table can be found in Appendix~\ref{sec:appendix_amb_rec}.}
\label{tab:ad_main_results}
\end{table*}

\subsection{Dialogue Intervention Analysis}

We study which properties of the dialogue history support grounding through controlled input interventions on Qwen3-VL. Keeping the image and final ambiguous instruction fixed, we replace the Phase-2 dialogue with counterfactual variants.

We define twelve interventions across four dimensions. \textit{Reference conditions} include the original Phase-2 input (``Full dialogue'') and the Phase-1 instruction (human order) without dialogue (``Order only''). \textit{Order perturbations} preserve turns while disrupting sequence through random shuffling over five seeds (``Shuffle turns'') or full reversal (``Reverse order''). \textit{Content perturbations} separate informativeness from token count using irrelevant in-domain dialogue (``Irrelevant substitution''), semantically neutral acknowledgements (``Neutral substitution''), or augmented histories in which irrelevant filler is added at the end (``Irrelevant suffix'') or between turns (``Irrelevant interleaving''). \textit{Selective exposure} retains only subsets of the dialogue: the first or last $k{=}2$ turns (``First $K$ turns'', ``Last $K$ turns'') or speaker-specific transcripts (``Robot only'', ``Human only'').

We report the following metrics: mean Intersection over Union (mIoU), Acc@$0.5$, and $\Delta$mIoU relative to ``full dialogue'' performance.

\section{Results}
\label{sec:results}

We organise our analysis around three research questions: \textbf{RQ1:} Can current VLMs recognize multimodal ambiguity, and does the difficulty vary between visual and linguistic categories? \textbf{RQ2:} Does dialogue history improve visual grounding under ambiguity, and to what extent does the benefit differ across ambiguity types? \textbf{RQ3:} Does dialogue content contribute to visual grounding under ambiguity beyond surface properties such as length and format?

Our experiments reveal a consistent asymmetry across both tasks: VLMs handle linguistic ambiguity substantially better than visual ambiguity. Additionally, dialogue provides the largest grounding gains precisely where it can supply the missing linguistic referent for pronoun and zero anaphora, leaving visual ambiguity as a persistent bottleneck.

\subsection*{RQ1: Multimodal Ambiguity Recognition}
\label{sec:results_recognition}

\paragraph{Overall performance.}
Table~\ref{tab:ad_main_results} reports macro-F1 and per-category recall for all models. Multilingual VLMs clearly outperform their Japanese-specific counterparts: Qwen3-VL and Gemma3 reach the highest overall scores (83.0 and 78.5 F1-M, respectively), whereas the best Japanese model, LLM-jp~VILA, achieves 60.1. The remaining Japanese models, Sarashina2 (40.3) and EvoVLM-JP (32.6), perform below the chance prediction level, underscoring a wide capability gap between model families on multimodal ambiguity recognition.

\paragraph{Visual and linguistic asymmetry.}
The per-category breakdown exposes a sharp split. Multilingual VLMs (Qwen3-VL and Gemma3) achieve near-ceiling recall on linguistic ambiguity: 96.7--100.0 on \textit{pronoun} and 93.1--100.0 on \textit{zero\_anaphora}. In contrast, performance collapses on visual categories, with \textit{same\_object} recall, for instance, dropping to 32.8 and 39.7 for these same models. This asymmetry directly answers \textbf{RQ1}: VLMs are strong at detecting underspecification in the text (pronouns, missing arguments) but largely fail to detect that multiple regions in the image satisfy the instruction constraints, where multimodal reasoning is actually required.

\begin{table*}[t]
\centering
\small
\begin{tabular}{llcccc}
\toprule
\textbf{Model} & \textbf{Phase} & \textbf{Acc@0.5 (\%)} & \textbf{Acc@0.75 (\%)} & \textbf{mIoU} & \textbf{\#Correct} \\
\midrule
GLIP & Instruction-Only & 26.9 & 24.4 & 0.272 & 129 \\
GLIP & Dialogue-Enhanced & 53.3 & 49.4 & 0.501 & 256 \\
\midrule
MDETR & Instruction-Only & 25.4 & 17.9 & 0.239 & 122 \\
MDETR & Dialogue-Enhanced & 54.0 & 32.9 & 0.449 & 259 \\
\midrule
Qwen3-VL & Instruction-Only & 63.5 & 56.5 & 0.585 & 305 \\
Qwen3-VL & Dialogue-Enhanced & \textbf{79.0} & \textbf{68.3} & \textbf{0.717} & \textbf{379} \\
\bottomrule
\end{tabular}
\caption{Visual grounding on our Japanese multimodal test set ($N{=}480$) on two evaluation phases. \textbf{Instruction-Only}: image plus the human instruction only. \textbf{Dialogue-Enhanced}: image plus dialogue-informed text. \textbf{\#Correct}: number of instances with a top-1 IoU $\geq 0.5$.}
\label{tab:main_results}
\end{table*}

\begin{table*}[t]
\centering
\small
\begin{tabular}{lccc}
\toprule
\textbf{Ambiguity Type} & \textbf{GLIP} & \textbf{MDETR} & \textbf{Qwen3-VL} \\
\midrule
Non-ambiguous & $61.7\,(\textcolor{ForestGreen}{\uparrow +21.8})$ & $63.3\,(\textcolor{ForestGreen}{\uparrow +24.2})$ & $87.9\,(\textcolor{ForestGreen}{\uparrow +0.4})$ \\
Ambiguous & $44.4\,(\textcolor{ForestGreen}{\uparrow +31.5})$ & $44.0\,(\textcolor{ForestGreen}{\uparrow +33.2})$ & $69.4\,(\textcolor{ForestGreen}{\uparrow +31.5})$ \\
\midrule
same\_object & $32.8\,(\textcolor{ForestGreen}{\uparrow +12.1})$ & $25.9\,(\textcolor{ForestGreen}{\uparrow +12.1})$ & $62.1\,(\textcolor{ForestGreen}{\uparrow +6.9})$ \\
same\_attribute & $38.2\,(\textcolor{ForestGreen}{\uparrow +18.2})$ & $38.2\,(\textcolor{ForestGreen}{\uparrow +14.5})$ & $67.3\,(\textcolor{ForestGreen}{\uparrow +14.5})$ \\
pronoun & $52.5\,(\textcolor{ForestGreen}{\uparrow +44.3})$ & $54.1\,(\textcolor{ForestGreen}{\uparrow +50.8})$ & $63.9\,(\textcolor{ForestGreen}{\uparrow +42.6})$ \\
zero\_anaphora & $53.4\,(\textcolor{ForestGreen}{\uparrow +50.0})$ & $56.9\,(\textcolor{ForestGreen}{\uparrow +53.4})$ & $84.5\,(\textcolor{ForestGreen}{\uparrow +60.3})$ \\
\bottomrule
\end{tabular}
\caption{Acc@0.5 (\%) under \textbf{Dialogue-Enhanced} grounding, broken down by dataset slice (non-ambiguous vs.\ ambiguous instructions, and the four ambiguity categories among ambiguous ones). Values in parentheses give the absolute gain over \textbf{Instruction-Only} on the same slice ($\Delta$Acc@0.5).}
\label{tab:ambiguity_type_improvements}
\end{table*}

\paragraph{Degenerate baselines.}
Two Japanese-specific VLMs exhibit near-degenerate decision behaviour: EvoVLM-JP predicts \textit{ambiguous} for every input (0.0 non-ambiguous recall, 100.0 across all positive categories), and Sarashina2 oscillates between the two extremes across prompt variants. High per-category recall values for these models are therefore artefacts of class bias rather than genuine understanding.

\subsection*{RQ2: Visual Grounding under Ambiguity}
\label{sec:results_grounding}

\paragraph{Dialogue-enhanced grounding improvements.}
Table~\ref{tab:main_results} shows that adding dialogue context substantially improves all three grounding models. For GLIP and MDETR, Acc@0.5 roughly doubles (26.9$\to$53.3 and 25.4$\to$54.0, respectively) and mean IoU
increases by 0.229 and 0.210 points. Qwen3-VL, the strongest model overall, still gains 15.5 R@1$_{0.5}$ points (63.5$\to$79.0) and 0.132 IoU points when dialogue is available.

An important methodological distinction underlies these results. For GLIP and MDETR, the ``Dialogue-Enhanced'' condition passes dialogue through a GPT-5.2 integration step that condensates the conversational turns into a single sentence before grounding. The reported gains, therefore, reflect the performance of a pipeline (grounding model + LLM integrator), not the grounding model's intrinsic ability to use dialogue. Qwen3-VL, by contrast, receives the full conversation history directly and shows stronger evidence that a VLM can itself exploit conversational context for grounding.

\paragraph{Grounding visual–linguistic asymmetry.}
Table~\ref{tab:ambiguity_type_improvements} disaggregates
accuracy gains by ambiguity type and reveals a pattern that mirrors the recognition results. Linguistic categories benefit dramatically: zero-anaphora gains reach $+50.0$, $+53.4$, and $+60.3$ Acc@0.5 for GLIP, MDETR, and Qwen, respectively, consistent with the hypothesis that dialogue context restores the missing argument needed to uniquely identify the referent. Visual categories remain low even after dialogue integration, proving that the dialogue enhancement does not eliminate the visual ambiguity gap; rather, it reallocates improvements to cases where linguistic clarification can resolve previously underspecified references.

\paragraph{Unambiguous instructions benefits.}
In the non-ambiguous condition, dialogue still produces modest gains (+21.8 and +24.2 for GLIP/MDETR; +0.4 for Qwen, where the instruction-only baseline is already high at 87.5). This suggests that even grounding-sufficient instructions benefit from the additional relational and spatial context supplied by prior turns.

\subsection{RQ3: Is Dialogue Content Causally Responsible?}
\label{sec:results_ablation}

\begin{figure}[t]
\centering
\includegraphics[width=\linewidth]{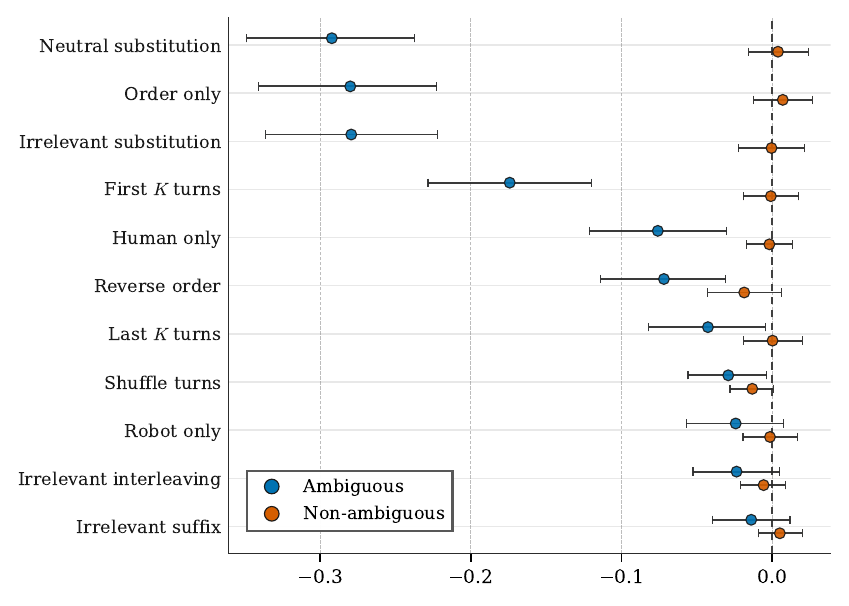}
\caption{Change in grounding performance under targeted interventions on the dialogue history for Qwen. Points show the mean per-sample $\Delta$IoU vs.\ the full-dialogue baseline; horizontal bars are 95\% confidence intervals across samples. Ambiguous instances (blue) are substantially more sensitive to interventions than non-ambiguous instances (orange). Extended results appear in Appendix~\ref{sec:appendix_dialogue_intervention}.}
\label{fig:dialogue_intervention}
\end{figure}

\paragraph{Content, not format, drives the gains.}
Figure~\ref{fig:dialogue_intervention} reports results for twelve targeted dialogue interventions on Qwen3-VL. The most informative comparison is between the content-removal conditions, such as \textit{neutral substitution} ($\Delta\text{IoU} = -0.292$) and \textit{irrelevant substitution} ($-0.279$), all of which reduce ambiguous performance to near the instruction-only baseline ($-0.280$) despite preserving dialogue length and structure. 

Additionally, structural perturbations that preserve content while disrupting ordering (shuffled turns: $-0.029$; reversed turns: $-0.072$) introduce substantially smaller penalties than content removal. Similarly, retaining only the final $K$ turns ($-0.043$) preserves most of the benefit, while retaining only the first $K$ ($-0.174$) is more costly, thus consistent with an expectation that the most recent referential context is most informative for grounding. These results directly answers \textbf{RQ3}: the grounding improvements are attributable to the semantic content of dialogue, not to any formatting or structural artefact.

\paragraph{Non-ambiguous instances robustness.}
Across every intervention, $\Delta\text{IoU}$ for non-ambiguous instances remains near zero (range: $-0.019$ to $+0.007$). This robustness is expected: when the instruction is already sufficient to uniquely identify the referent, dialogue carries at most redundant information, and its removal or perturbation has no systematic effect. The sharp contrast with ambiguous instances ($\Delta\text{IoU}$ up to $-0.292$) provides strong evidence that the causal role of dialogue is specifically to resolve multimodal ambiguity rather than to supply generic perceptual context.

\begin{figure}[t]
    \centering
    \includegraphics[width=\linewidth]{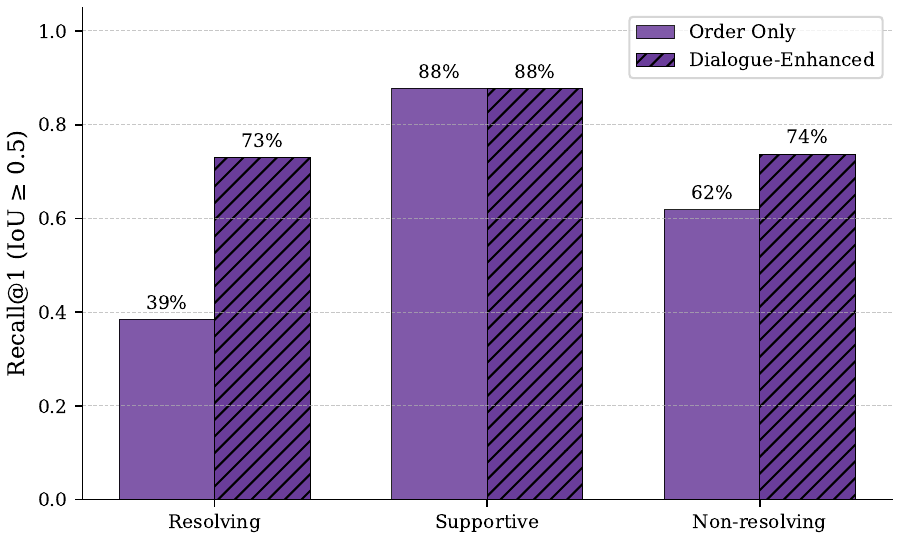}
    \caption{Qwen3-VL grounding performance (Acc@0.5) comparing \textit{Instruction-Only} vs \textit{Dialogue-Enhanced} across dialogue resolution modes: \textit{Resolving} (dialogue disambiguates the referent), \textit{Supportive} (order is already clear; dialogue adds context), and \textit{Non-resolving} (dialogue does not disambiguate).}
    \label{fig:dialogue_mode}
\end{figure}

\paragraph{Non-resolving dialogue modest benefits.}
Regarding the dialogue-mode analysis (Figure~\ref{fig:dialogue_mode}), the \textit{resolving} condition yields the largest absolute gain (39$\to$73\% accuracy), as expected. Interestingly, the \textit{non-resolving} condition still improves over instruction-only (62$\to$74\%), despite dialogue that was designed not to disambiguate. The intervention results suggest a partial explanation: dialogue may supply auxiliary grounding cues (spatial relations, object attributes) that are useful even when they do not narrow the set of valid referents.

\section{Conclusion}

We present a systematic analysis of ambiguity in visual grounding through JAG-D, a new Japanese test set that evaluates visual and linguistic ambiguity in human-robot interaction scenarios, providing dialogue as a disambiguation context. Across both ambiguity recognition and grounding, we observe a consistent asymmetry: Japanese and multilingual VLMs resolve discourse-driven (linguistic) ambiguity substantially better than scene-driven (visual) ambiguity. We show that current VLMs can recognize linguistic ambiguity with high confidence (93.1-100.0 recall), while performing poorly on visual ambiguity (32.8-54.5 recall). For visual grounding, dialogue context substantially improves performance for linguistic cases (up to $+60.3$ Acc@0.5 for zero-anaphora), whereas visual ambiguity remains a persistent bottleneck, with 25.9 accuracy in the worst case even after dialogue enhancement. Overall, our findings identify visual ambiguity as a central unresolved challenge for robust conversational grounding and motivate models capable of reasoning over multiple modalities beyond surface-level linguistic pattern matching.

\section*{Limitations}

First, our benchmark size is relatively small compared to standard grounding datasets. However, it is explicitly designed as a highly curated diagnostic test set requiring specific real-world objects and precise multi-turn dialogue. We prioritized linguistic fidelity via native Japanese annotator validation to ensure the accurate representation of complex and context-dependent phenomena like zero anaphora. Second, our findings are currently limited to Japanese; future work must explore whether this visual-linguistic asymmetry persists in a range of distinct languages. Finally, our dialogue-enhanced evaluation for GLIP and MDETR relies on an LLM-based (GPT-5.2) text integration pipeline. This was necessitated by the current lack of Japanese-specific visual grounding models capable of natively processing long-context dialogue histories. Consequently, reported gains for these architectures reflect the pipeline's effectiveness rather than the grounding models' intrinsic multi-turn capabilities.

\section*{Ethical Considerations}

Our test set utilizes real-world human-robot interaction images from the J-CRe3 dataset \cite{ueda-etal-2024-j}.  In building upon this dataset, we ensured that our newly generated multi-turn dialogues and annotations did not introduce any sensitive or personally identifiable information regarding the actors in the visual sources. Furthermore, all manual dataset validation was conducted by native Japanese annotators who were compensated according to the task. Lastly, deploying multimodal models in physical robotic systems carries inherent safety risks if ambiguous commands are misinterpreted. By exposing current VLM failure modes regarding perceptual ambiguity, we aim to encourage the development of safer agents that actively seek clarification rather than executing potentially harmful actions based on underspecified instructions.

\section*{Acknowledgments}

We would like to thank Hideko Habe and Yoko Matsui for their valuable contributions to the annotation of our benchmark. A part of this work was supported by JST, PRESTO Grant Number JPMJPR24TC, Japan.

\bibliography{main.bib}

\appendix

\section{JAG-D Benchmark}
\label{sec:appendix_data}

\subsection{J-CRe3 Images}
\label{sec:appendix_jcre3_images}
\begin{figure*}[htbp]
    \centering
    \begin{subfigure}[b]{0.3\textwidth}
        \centering
        \includegraphics[width=\textwidth]{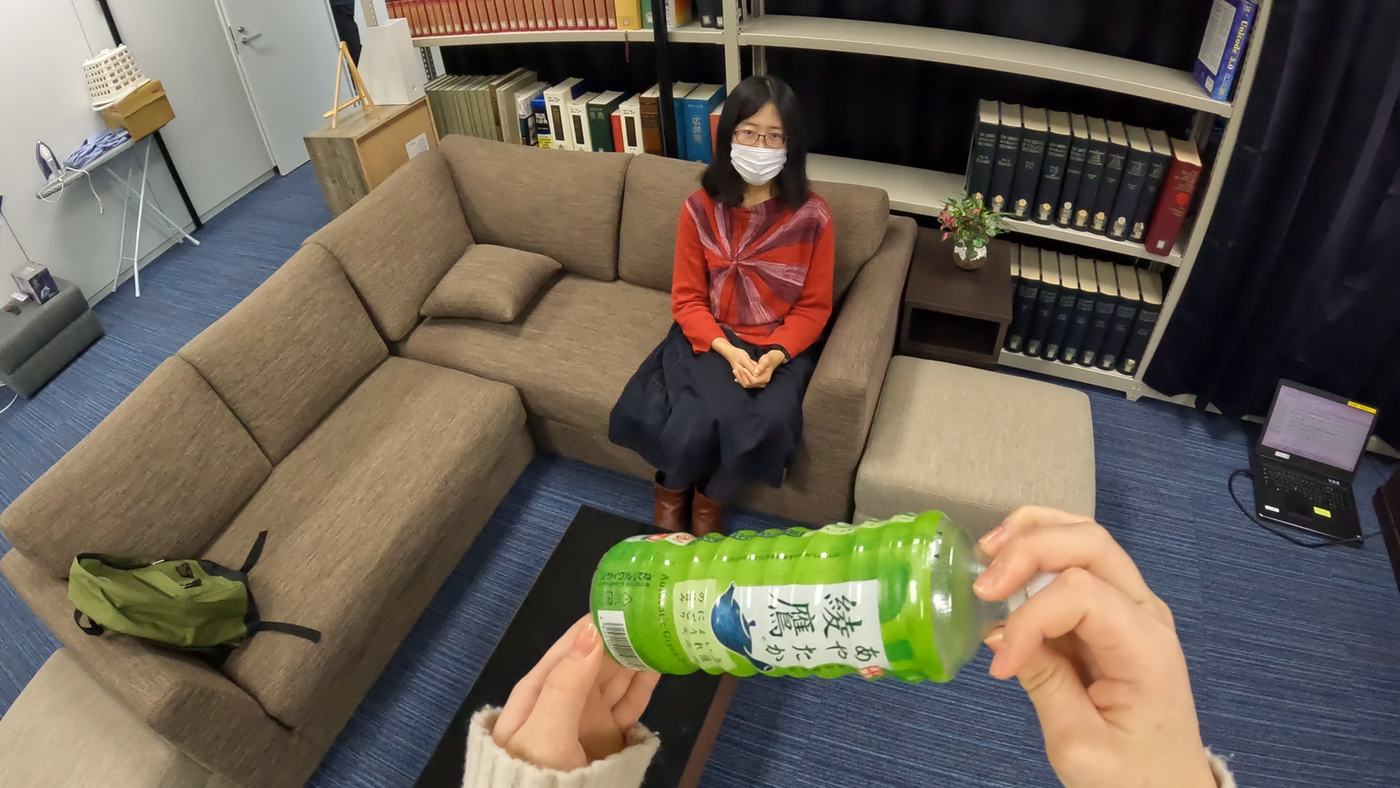}
        \caption{Living room}
        \label{fig:img1}
    \end{subfigure}
    \hfill 
    \begin{subfigure}[b]{0.3\textwidth}
        \centering
        \includegraphics[width=\textwidth]{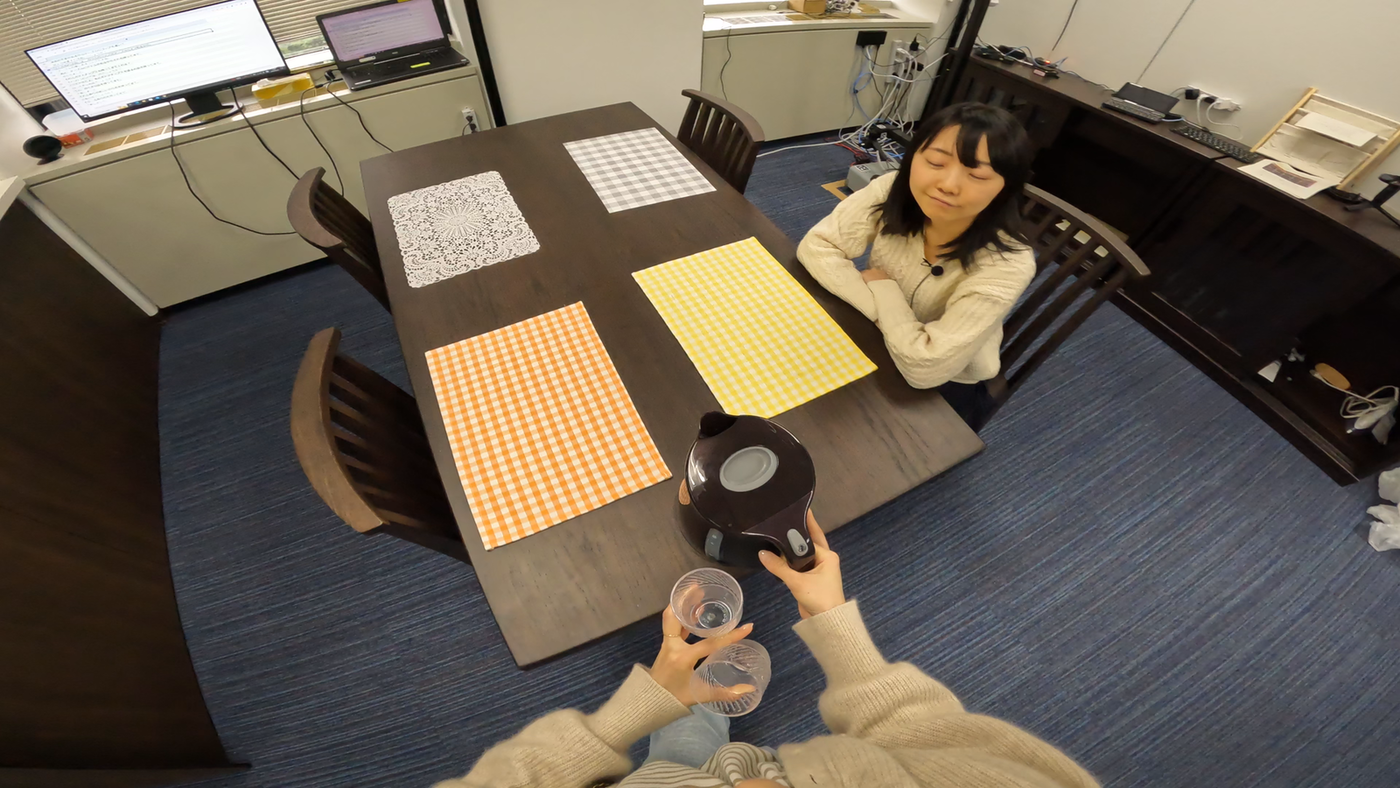}
        \caption{Dining room}
        \label{fig:img2}
    \end{subfigure}
    \hfill 
    \begin{subfigure}[b]{0.3\textwidth}
        \centering
        \includegraphics[width=\textwidth]{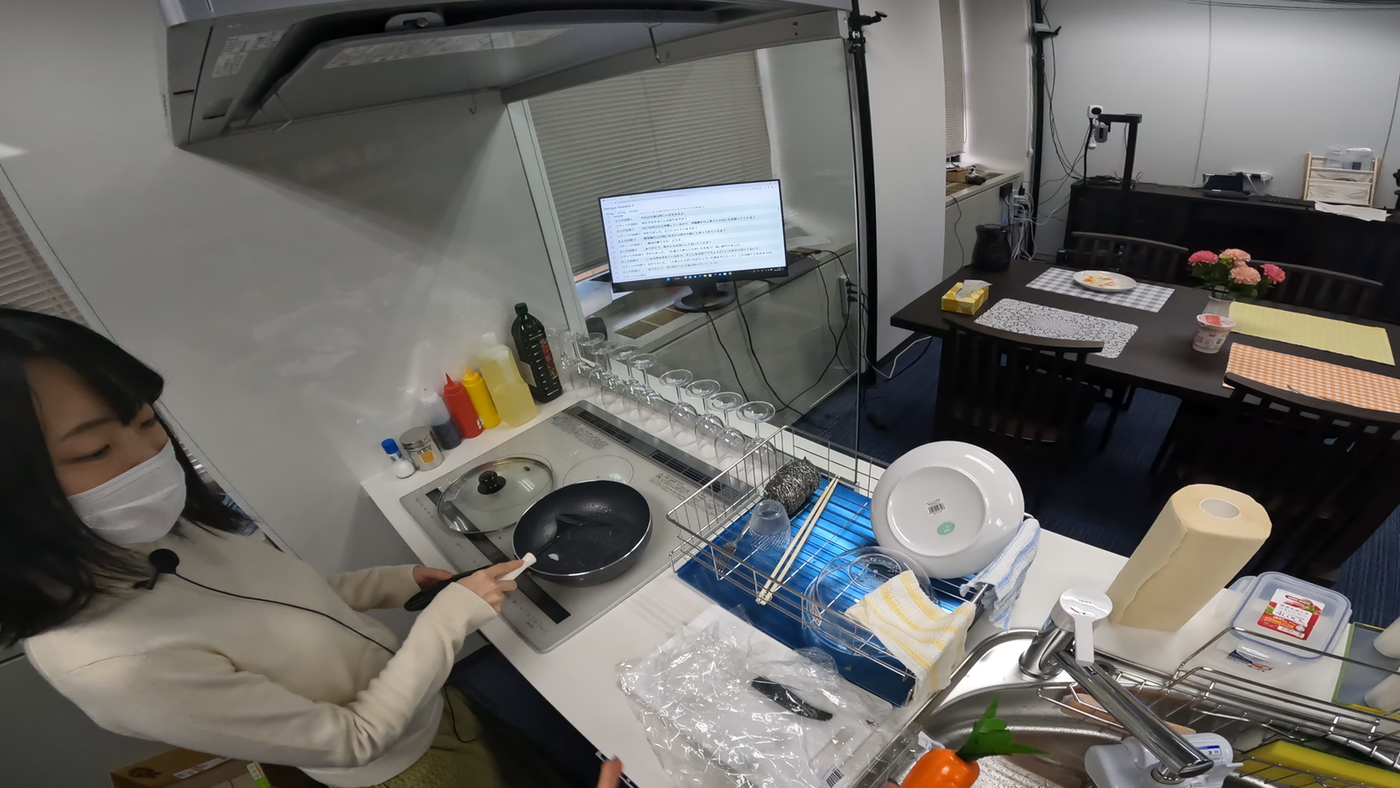}
        \caption{Kitchen}
        \label{fig:img3}
    \end{subfigure}
    
    \caption{J-CRe3 scenario examples}
    \label{fig:jcre3_images}
\end{figure*}

Our benchmark's images are sampled from J-CRe3 \citep{ueda-etal-2024-j}. J-CRe3 contains egocentric human-robot interaction scenarios in real-world environments. These environments represent different elements of rooms where interactions with social robots usually take place, such as the living room, dining room, and kitchen (Figure~\ref{fig:jcre3_images}).

\subsection{Dialogue Mode Examples}
\label{sec:appendix_dialogue_mode}

In our benchmark, we divide dialogue history data into three specific modes:
\begin{itemize}
    \item \textbf{Resolving}: the dialogue content helps in disambiguating the ambiguous human instruction.
    \item \textbf{Non-resolving}: the dialogue doesn't contain any relevant information for the current instruction (either ambiguous or not).
    \item \textbf{Supportive}: the dialogue contains relevant information for the current unambiguous instruction. The last is already clear, but still, the dialogue can help the grounding model with additional context.
\end{itemize}

Table~\ref{tab:example_dialogue_mode} contains examples for every introduced dialogue mode.

\renewcommand{\arraystretch}{1.4}
\begin{table*}[ht]
\centering
\small
\begin{tabular}{
>{\raggedright\arraybackslash}m{2cm}
>{\centering\arraybackslash}m{4cm}
>{\centering\arraybackslash}m{6cm}
}
\toprule
\textbf{Dialogue Mode} & \textbf{Final Order} & \textbf{Dialogue History} \\
\midrule

\textit{\textbf{resolving}} & 
\textit{``Hand me the rectangular item.''} & 
\parbox[t]{6cm}{
``robot'': \textit{``I see a green backpack on the couch. Is there something you need from it?''} \\
``master'': \textit{``Yes, there's an item inside the main compartment.''} \\
``robot'': \textit{``I notice a \textcolor{ForestGreen}{black electronic device} between your hands as you adjust the zipper. Is that the item?''} \\
``master'': \textit{``Exactly, that's the one I need.''} \\
``robot'': \textit{``It seems to be \textcolor{ForestGreen}{rectangular} and could be \textcolor{ForestGreen}{a tablet or a laptop}. Should I retrieve it for you?''} \\
``master'': \textit{``Yes, please hand me the \textcolor{ForestGreen}{rectangular item}.''}
}
\\
\midrule

\textit{\textbf{non-resolving}} & 
\textit{``Hand me the mug.''} & 
\parbox[t]{6cm}{
``robot'': ``I noticed there are \textcolor{BrickRed}{two white mugs} on the coffee table.'' \\
``master'': \textit{``Yes, \textcolor{BrickRed}{they both look quite similar}, don't they?''} \\
``robot'': \textit{``Indeed. \textcolor{BrickRed}{One is} sitting right next to the yellow tissue box.''} \\
``master'': \textit{``And \textcolor{BrickRed}{the other is} right beside it, in front of me.''} \\
``robot'': \textit{``They're both conveniently placed on the black coffee table.''} \\
``master'': \textit{``Could you hand me \textcolor{BrickRed}{the mug}?''}
}
\\
\midrule

\textit{\textbf{supportive}} & 
\textit{``Hand me the notebook.''} &
\parbox[t]{6cm}{
``robot'': \textit{``I'm here to assist. Is there something specific you need today?''} \\
``master'': \textit{``Yes, I need \textcolor{ForestGreen}{the notebook}. It's on the couch.''} \\
``robot'': \textit{``I see multiple items. Is it near anything in particular?''} \\
``master'': \textit{``The notebook is \textcolor{ForestGreen}{dark blue} and should be \textcolor{ForestGreen}{to the left of the green backpack}.''} \\
``robot'': \textit{``Ah, and is it \textcolor{ForestGreen}{under a piece of white paper with Japanese text}?''} \\
``master'': \textit{``Yes, that's the one. Hand me \textcolor{ForestGreen}{the notebook}.''}
}
\\

\bottomrule
\end{tabular}
\caption{Example of dialogue mode in JAG-D. The Japanese version is omitted due to space constraints. Note that the real data is in Japanese.}
\label{tab:example_dialogue_mode}
\end{table*}

\subsection{Data Generation Pipeline}
\label{sec:appendix_data_generation}
\begin{figure*}[h]
\centering
\small
\resizebox{1.0\textwidth}{!}{
\begin{tikzpicture}[
    model/.style={
        draw, rounded corners, align=center,
        minimum width=3.5cm, minimum height=1.8cm,
        fill=gray!10, font=\sffamily\normalsize
    },
    dialogue/.style={
        draw, dashed, align=center,
        minimum width=2.8cm, minimum height=1.5cm,
        fill=white, font=\sffamily\normalsize
    },
    arrow/.style={-{Stealth[length=3mm,width=2mm]}, thick, draw=black!70},
    darrow/.style={-{Stealth[length=3mm,width=2mm]}, thick, dashed, draw=black!70}
]

\node (input) at (-6.5, -0.5) {
    \includegraphics[width=3.5cm]{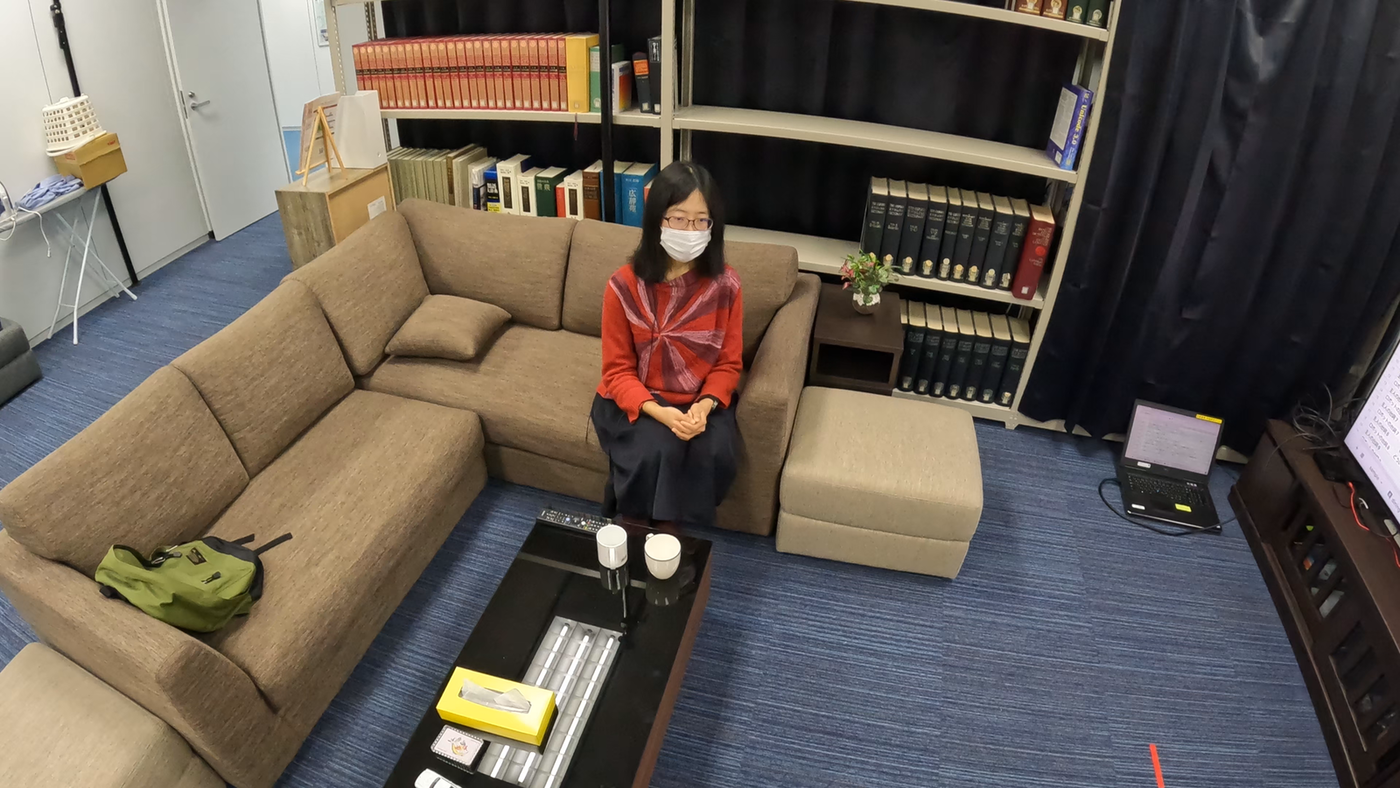}
};

\node[model] (qwentop) at (0, 3) {
    \includegraphics[height=0.8cm]{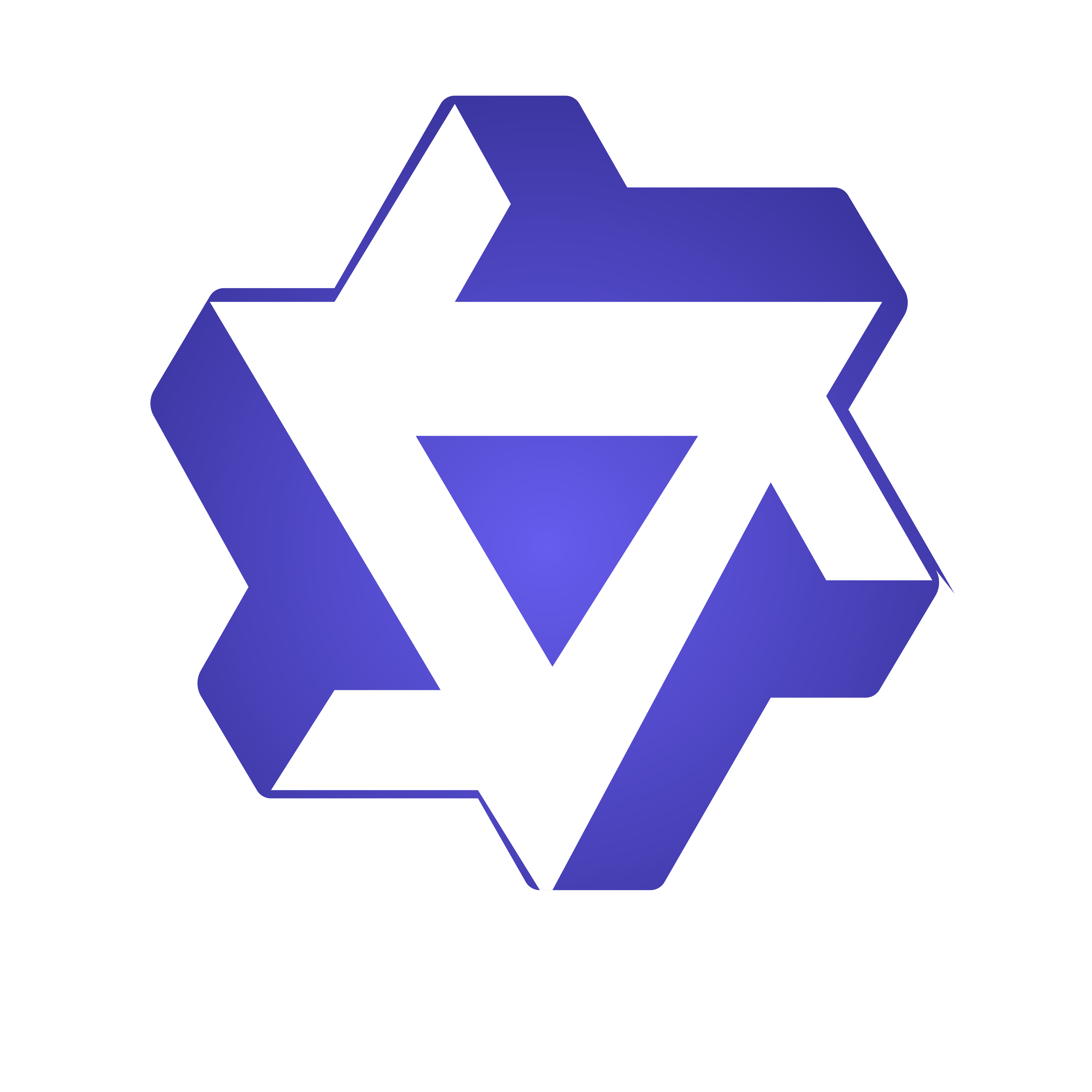} \\[3pt]
    Qwen3-VL
};
\node[model] (gptbot)  at (0, -3) {
    \includegraphics[height=0.8cm]{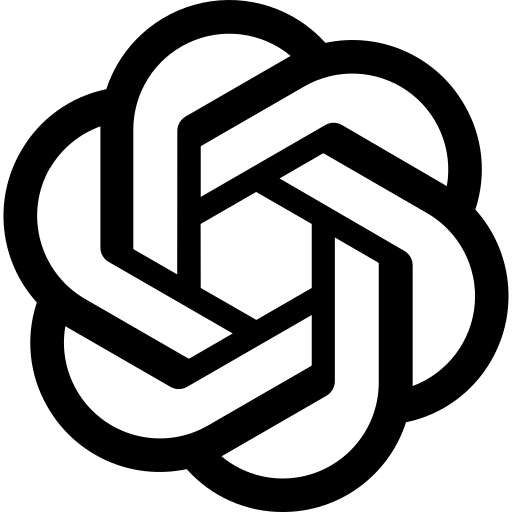} \\[3pt]
    GPT-5.1
};

\node[model] (gpttop)  at (5, 3) {
    \includegraphics[height=0.8cm]{figures/icons/openai_logo.png} \\[3pt]
    GPT-5.1
};
\node[model] (qwenmid) at (5, -0.5) {
    \includegraphics[height=0.8cm]{figures/icons/qwen_logo.png} \\[3pt]
    Qwen3-VL
};
\node[model] (dino)    at (5, -3) {
    \includegraphics[height=0.8cm]{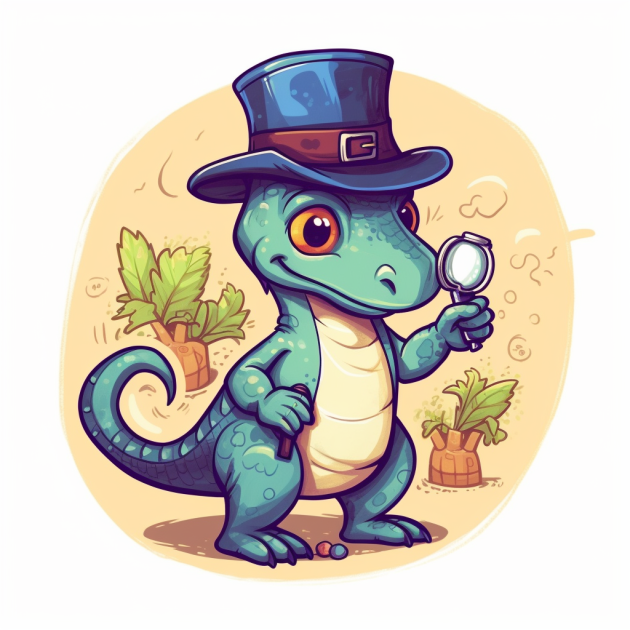} \\[3pt]
    Grounding DINO
};

\node[model] (gptright) at (10, 1.25) {
    \includegraphics[height=0.8cm]{figures/icons/openai_logo.png} \\[3pt]
    GPT-5.1
};
\node[dialogue] (dial)  at (10, -1.25) {Dialogue\\History};

\node[dialogue] (output) at (10, -3) {Bounding\\Boxes};

\draw[darrow] (input.east) -- (qwentop.west);
\draw[darrow] (input.east) -- (qwenmid.west);

\draw[arrow] (qwentop.east) -- (gpttop.west) node[midway, above, font=\sffamily\small, text=black!80] {Caption};

\draw[arrow] (qwentop.south) -- (gptbot.north) node[midway, right, font=\sffamily\small, text=black!80, align=center] {Image\\Caption};

\draw[arrow] ([xshift=-0.8cm]gpttop.south) -- ([xshift=-0.8cm]qwenmid.north) node[midway, right, font=\sffamily\small, text=black!80, align=center] {Ambiguity\\Instruction};

\draw[arrow] (gptbot.east) -- (dino.west) node[midway, above, font=\sffamily\small, text=black!80] {Object Extracted};

\draw[arrow] (gpttop.east) -- (gptright.west) node[pos=0.4, above right, xshift=-0.2cm, font=\sffamily\small, text=black!80, align=center] {Ambiguity\\Instruction};
\draw[arrow] (qwenmid.east) -- (gptright.west) node[pos=0.4, below right, xshift=-0.2cm, font=\sffamily\small, text=black!80, align=center] {Target\\Object};

\draw[darrow] (gptright.south) -- (dial.north);

\draw[darrow] (dino.east) -- (output.west);

\end{tikzpicture}
}
\caption{JAG-D data generation pipeline combining Qwen3-VL, GPT-5.1, and Grounding DINO.}
\label{fig:appendix_data_pipeline}
\end{figure*}

Starting from J-CRe3, the remaining data for JAG-D has been synthetically generated using a combination of three models: Qwen3-VL\footnote{\url{https://huggingface.co/Qwen/Qwen3-VL-8B-Instruct}}, GPT-5.1\footnote{\url{https://developers.openai.com/api/docs/models/gpt-5.1}}, and Grounding DINO\footnote{\url{https://huggingface.co/IDEA-Research/grounding-dino-base}}. Figure~\ref{fig:appendix_data_pipeline} represents the data generation pipeline for our benchmark. Specifically, Qwen3-VL receive the social scenario image as input and is prompted to generate a detailed image caption. From this caption, GPT-5.1 is prompted to generate complex textual data: the ambiguous/non-ambiguous instruction and the related dialogue history. GPT-5.1 and Grounding DINO are also used to generate the bounding box candidates for the visual grounding task. The synthetic data is then human-curated and collected by Japanese native professional annotators to assess the data quality of our designed benchmark. All prompts used in the pipeline can be found in our code repository.

\section{Ambiguity Recognition Experiment}
\label{sec:appendix_amb_rec}

We evaluate Japanese and multilingual VLMs on the ambiguity recognition task. Figure~\ref{fig:appendix_zs_amb_rec} and \ref{fig:appendix_fs_amb_rec} contain the zero-shot and few-shot prompts for these models. We evaluate them under the two prompt settings, in Japanese and in English, for multilingual models only. Table~\ref{tab:appendix_amb_rec_overall} reports the extended results for the VLMs tested. In addition, Table~\ref{tab:appendix_amb_rec_category_performance} reports fine-grained performance, comparing different categories of ambiguity.

\begin{figure*}[t]
  \begin{tcolorbox}[
    colback=white,
    colframe=black!70,
    title={\small\textbf{Zero-shot Ambiguity Recognition Prompt}},
    coltitle=white,
    colbacktitle=black!75,
    boxrule=1pt,
    arc=1pt,
    titlerule=0pt,
    toptitle=2pt,
    bottomtitle=2pt,
    left=6pt, right=6pt, top=6pt, bottom=6pt
  ]
  \small
Look at the image and the instructions below. Determine whether the instruction is ambiguous or not.

\vspace{4pt}
An instruction is considered "ambiguous" if the mentioned object is not clearly identifiable in the image.
An instruction is considered "non-ambiguous" if the mentioned object is clearly identifiable in the image.

\vspace{4pt}
Reply with ONLY 'ambiguous' or 'non-ambiguous'.

\vspace{4pt}
Instruction: \{\{instruction\}\}
  \end{tcolorbox}
\caption{Zero-shot ambiguity recognition prompt. The Japanese version is the same translated prompt.}
\label{fig:appendix_zs_amb_rec}
\end{figure*}

\begin{figure*}[t]
  \begin{tcolorbox}[
    colback=white,
    colframe=black!70,
    title={\small\textbf{Few-shot Ambiguity Recognition Prompt}},
    coltitle=white,
    colbacktitle=black!75,
    boxrule=1pt,
    arc=1pt,
    titlerule=0pt,
    toptitle=2pt,
    bottomtitle=2pt,
    left=6pt, right=6pt, top=6pt, bottom=6pt
  ]
  \small
Look at the image and the instructions below. Determine whether the instruction is ambiguous or not.

\vspace{4pt}
An instruction is considered "ambiguous" if the mentioned object is not clearly identifiable in the image.
An instruction is considered "non-ambiguous" if the mentioned object is clearly identifiable in the image.

\vspace{4pt}
\{\{EXAMPLES\_EN\}\}

\vspace{4pt}
---

\vspace{4pt}
[Target for Classification]

Instruction: "\{\{instruction\}\}"

Scenario: (See the input image)

\vspace{4pt}
Reply with ONLY 'ambiguous' or 'non-ambiguous'.
Classification: 
  \end{tcolorbox}
\caption{Few-shot ambiguity recognition prompt. The Japanese version is the same translated prompt.}
\label{fig:appendix_fs_amb_rec}
\end{figure*}

\begin{table*}[h]
\centering
\small
\begin{tabular}{llcccccc}
\toprule
\textbf{Model} & \textbf{Prompt} & \textbf{Lang} & \textbf{Acc (\%)} & \textbf{Prec (\%)} & \textbf{Rec (\%)} & \textbf{F1 (\%)} & \textbf{F1-M (\%)} \\
\midrule
Qwen3-VL & Few-shot & EN & 83.3 & 92.2 & 71.6 & 80.6 & 83.0 \\
Qwen3-VL & Few-shot & JA & 82.5 & 92.5 & 69.4 & 79.3 & 82.1 \\
Gemma3 & Few-shot & EN & 78.8 & 82.5 & 71.1 & 76.4 & 78.5 \\
Qwen3-VL & Zero-shot & JA & 76.9 & 79.5 & 70.3 & 74.6 & 76.7 \\
Gemma3 & Few-shot & JA & 76.5 & 70.4 & 88.4 & 78.4 & 76.3 \\
Gemma3 & Zero-shot & JA & 74.2 & 77.8 & 65.1 & 70.9 & 73.8 \\
Qwen3-VL & Zero-shot & EN & 69.8 & 92.2 & 40.9 & 56.7 & 66.8 \\
Gemma3 & Zero-shot & EN & 61.5 & 69.4 & 36.2 & 47.6 & 58.6 \\
LLM-jp VILA & Few-shot & JA & 60.4 & 60.3 & 53.0 & 56.4 & 60.1 \\
LLM-jp VILA & Zero-shot & JA & 56.7 & 57.0 & 42.2 & 48.5 & 55.6 \\
Sarashina2 & Zero-shot & JA & 52.1 & 100.0 & 0.9 & 1.7 & 35.0 \\
Sarashina2 & Few-shot & JA & 51.9 & 50.1 & 99.1 & 66.6 & 40.3 \\
EvoVLM-JP & Zero-shot & JA & 48.3 & 48.3 & 100.0 & 65.2 & 32.6 \\
EvoVLM-JP & Few-shot & JA & 48.3 & 48.3 & 100.0 & 65.2 & 32.6 \\
\bottomrule
\end{tabular}
\caption{Multimodal ambiguity recognition with full classification metrics on the complete test set ($N{=}480$): accuracy, precision, recall, micro-F1, and macro-F1 (F1-M), spanning prompt protocol (few-shot vs.\ zero-shot) and language for multilingual VLMs. Rows are sorted by accuracy (descending).}
\label{tab:appendix_amb_rec_overall}
\end{table*}

\begin{table*}[h]
\centering
\small
\begin{tabular}{ll|c|cc|cc}
\toprule
& & \textbf{Non-Ambig} & \multicolumn{2}{c|}{\textbf{Visual Ambiguity}} & \multicolumn{2}{c}{\textbf{Linguistic Ambiguity}} \\
\textbf{Model} & \textbf{Config} & \textbf{Recall} & \textbf{Same Obj} & \textbf{Same Attr} & \textbf{Pronoun} & \textbf{Zero Ana} \\
\midrule
Qwen3-VL & FS-EN & 94.4 & 32.8 & 54.5 & 96.7 & 100.0 \\
Qwen3-VL & FS-JA & 94.8 & 32.8 & 52.7 & 95.1 & 94.8 \\
Qwen3-VL & ZS-JA & 83.1 & 56.9 & 41.8 & 86.9 & 93.1 \\
Qwen3-VL & ZS-EN & 96.8 & 6.9 & 14.5 & 72.1 & 67.2 \\
\midrule
Gemma3 & FS-EN & 85.9 & 39.7 & 52.7 & 96.7 & 93.1 \\
Gemma3 & FS-JA & 65.3 & 77.6 & 76.4 & 100.0 & 98.3 \\
Gemma3 & ZS-JA & 82.7 & 48.3 & 38.2 & 90.2 & 81.0 \\
Gemma3 & ZS-EN & 85.1 & 20.7 & 27.3 & 47.5 & 48.3 \\
\midrule
LLM-jp VILA & FS-JA & 67.3 & 46.6 & 54.5 & 47.5 & 63.8 \\
LLM-jp VILA & ZS-JA & 70.2 & 41.4 & 32.7 & 41.0 & 53.4 \\
\midrule
Sarashina2 & FS-JA & 7.7 & 98.3 & 98.2 & 100.0 & 100.0 \\
Sarashina2 & ZS-JA & 100.0 & 1.7 & 0.0 & 0.0 & 1.7 \\
\midrule
EvoVLM-JP & FS-JA & 0.0 & 100.0 & 100.0 & 100.0 & 100.0 \\
EvoVLM-JP & ZS-JA & 0.0 & 100.0 & 100.0 & 100.0 & 100.0 \\
\bottomrule
\end{tabular}
\caption{Fine-grained recall (\%) for multimodal ambiguity recognition: Non-Ambig reports recall on gold-unambiguous instances; visual (\textit{same\_object}, \textit{same\_attribute}) vs.\ linguistic (\textit{pronoun}, \textit{zero\_anaphora}) columns condition on gold ambiguity type among ambiguous instances. Each row is one fully specified model configuration (few-shot/zero-shot and English/Japanese).}
\label{tab:appendix_amb_rec_category_performance}
\end{table*}

\section{Visual Grounding Experiment}
\label{sec:appendix_grounding}

The visual grounding experiment compares two grounding models (GLIP and MDETR) and Qwen3-VL as a multilingual VLM. The models are tested in two different phases:
\begin{enumerate}
    \item \textbf{Instruction-only:} the models receive as input only the human instruction, which can be ambiguous or not.
    \item \textbf{Dialogue-enhanced:} in addition to the previous setting, the models receive a dialogue history as additional context, to potentially disambiguate the human instruction.
\end{enumerate}

GLIP and MDETR can't handle long context, such as the dialogue history. For this reason, we use a method from \citet{willemsen-etal-2023-resolving} to integrate the dialogue information into a single instruction (Figure~\ref{fig:appendix_grounding_pipeline}).

In contrast, Qwen3-VL handles long context, and we can pass it the dialogue history directly. Figure~\ref{fig:appendix_qwen_instruction_only} and Figure~\ref{fig:appendix_qwen_dialogue_enhanced} contains the instruction-only and the dialogue-enhanced prompts for Qwen, respectively.

\begin{figure*}[t]
  \begin{tcolorbox}[
    colback=white,
    colframe=black!70,
    title={\small\textbf{Grounding Dialogue Integration Prompt}},
    coltitle=white,
    colbacktitle=black!75,
    boxrule=1pt,
    arc=1pt,
    titlerule=0pt,
    toptitle=2pt,
    bottomtitle=2pt,
    left=6pt, right=6pt, top=6pt, bottom=6pt
  ]
  \small
You are an expert at converting ambiguous Japanese commands into precise object descriptions for visual grounding.

\#\# Context

Your output will be fed to a phrase grounding model that locates objects in images. 

\vspace{4pt}
CRITICAL: The model parses your output into noun phrases and grounds EACH ONE separately. If you write "The black laptop on the coffee table", the model will find:

- "Coffee table" → grounds to the table (WRONG!)

- "Laptop" → grounds to the laptop

And it may pick the table instead of the laptop!

\vspace{4pt}
\#\# Your Task

Output a SINGLE, SIMPLE noun phrase describing ONLY the target object with its INTRINSIC visual attributes.

\vspace{4pt}
\#\# Guidelines

1. Output ONLY the target object description — no verbs, no other objects, no context

2. DO NOT mention furniture, locations, or other objects

3. Include distinguishing visual attributes from the dialogue (color, size, shape, etc.)

4. Resolve pronouns (that, this, that) and zero anaphora using dialogue context

5. Keep it concise: ideally 2-6 words

6. Output in Japanese only, no explanations

\vspace{4pt}
\{\{EXAMPLES\_EN\}\}

\vspace{4pt}
Output ONLY the target object noun phrase:
  \end{tcolorbox}
\caption{GPT-5.2 dialogue integration prompt for the visual grounding experiment.}
\label{fig:appendix_grounding_pipeline}
\end{figure*}

\begin{figure*}[t]
  \begin{tcolorbox}[
    colback=white,
    colframe=black!70,
    title={\small\textbf{Qwen3-VL Instruction-only Prompt}},
    coltitle=white,
    colbacktitle=black!75,
    boxrule=1pt,
    arc=1pt,
    titlerule=0pt,
    toptitle=2pt,
    bottomtitle=2pt,
    left=6pt, right=6pt, top=6pt, bottom=6pt
  ]
  \small
You are an expert in visual grounding.

Given an image and a Japanese user instruction, identify the referred object and output its 2D bounding box.

\vspace{4pt}
\#\# Instruction

\{\{order\}\}

\vspace{4pt}
\#\# Output format (strict)

Output JSON only. No extra text.

Include these keys: \{``bbox\_2d'': $[x1,y1,x2,y2]$, ``score'': $0.0$\}

bbox\_2d must be normalized to $[0,1000]$ with origin at top-left: $[x1,y1,x2,y2]$.

score must be a confidence in [0.0, 1.0].
  \end{tcolorbox}
\caption{Qwen3-VL instruction-only prompt for the visual grounding experiment.}
\label{fig:appendix_qwen_instruction_only}
\end{figure*}

\begin{figure*}[t]
  \begin{tcolorbox}[
    colback=white,
    colframe=black!70,
    title={\small\textbf{Qwen3-VL Dialogue-enhanced Prompt}},
    coltitle=white,
    colbacktitle=black!75,
    boxrule=1pt,
    arc=1pt,
    titlerule=0pt,
    toptitle=2pt,
    bottomtitle=2pt,
    left=6pt, right=6pt, top=6pt, bottom=6pt
  ]
  \small
You are an expert in visual grounding.

Use the full Japanese dialogue history to resolve ambiguous references and identify the object referred to by the final instruction.

Ambiguity can arise because of multiple similar objects in the image or because of pronouns or omissions like zero anaphora in Japanese.

Use the dialogue history to identify the single target object.

\vspace{4pt}
\#\# Dialogue history
\{\{dialogue\_text\}\}

\vspace{4pt}
\#\# Final instruction
\{\{order\}\}

\vspace{4pt}
\#\# Output format (strict)

Output JSON only. No extra text.

Include these keys: \{``bbox\_2d'': $[x1,y1,x2,y2]$, ``score'': $0.0$\}

bbox\_2d must be normalized to $[0,1000]$ with origin at top-left: $[x1,y1,x2,y2]$.

score must be a confidence in $[0.0, 1.0]$.
  \end{tcolorbox}
\caption{Qwen3-VL dialogue-enhanced prompt for the visual grounding experiment.}
\label{fig:appendix_qwen_dialogue_enhanced}
\end{figure*}

\section{Dialogue Intervention Experiment}
\label{sec:appendix_dialogue_intervention}

To understand how the dialogue history impacts visual grounding in Qwen3-VL, we run controlled counterfactual interventions on the dialogue-enhanced grounding evaluation phase.
The interventions are the following:

\begin{itemize}
    \item \textbf{Full dialogue:} the original, unaltered dialogue history (Phase~2 baseline).
    \item \textbf{Order only:} the human instruction only, with no dialogue (Phase~1 baseline).
    \item \textbf{Shuffled turns:} the same dialogue turns in a random order.
    \item \textbf{Reversed turns:} the dialogue with its turn order reversed.
    \item \textbf{Irrelevant filler:} an unrelated dialogue, length-matched to the original to control for token volume.
    \item \textbf{Last $K$ turns:} only the final $K$ turns of the dialogue ($K{=}2$).
    \item \textbf{First $K$ turns:} only the initial $K$ turns of the dialogue ($K{=}2$).
    \item \textbf{Neutral substitution:} the dialogue is replaced with generic robot/human turns matched to the original length.
    \item \textbf{Irrelevant suffix:} the true dialogue followed by irrelevant turns.
    \item \textbf{Irrelevant interleaving:} the original dialogue with one irrelevant line inserted after each original turn.
    \item \textbf{Robot turns only:} the dialogue restricted to robot utterances.
    \item \textbf{Human turns only:} the dialogue restricted to human utterances.
\end{itemize}

The results of the intervention experiment are reported in Table~\ref{tab:appendix_qwen_dialogue_interventions}.

\begin{table*}[t]
\centering
\small
\begin{tabular}{l rrr rrr}
\toprule
\textbf{Intervention} & \multicolumn{3}{c}{\textbf{Ambiguous}} & \multicolumn{3}{c}{\textbf{Non-ambiguous}} \\
\cmidrule(lr){2-4} \cmidrule(lr){5-7}
& mIoU & $\Delta$IoU & Acc@$.5$ & mIoU & $\Delta$IoU & Acc@$.5$ \\
\midrule
Full dialogue (baseline) & 0.641 & 0.000 & 0.694 & 0.787 & 0.000 & 0.879 \\
Neutral substitution & 0.349 & -0.292 & 0.362 & 0.791 & 0.004 & 0.887 \\
Order only & 0.361 & -0.280 & 0.379 & 0.794 & 0.007 & 0.875 \\
Irrelevant substitution & 0.362 & -0.279 & 0.379 & 0.786 & -0.001 & 0.883 \\
First $K$ turns & 0.467 & -0.174 & 0.487 & 0.786 & -0.001 & 0.879 \\
Human only & 0.565 & -0.076 & 0.608 & 0.785 & -0.002 & 0.879 \\
Reversed turns & 0.569 & -0.072 & 0.608 & 0.768 & -0.019 & 0.843 \\
Last $K$ turns & 0.599 & -0.043 & 0.629 & 0.787 & 0.000 & 0.879 \\
Shuffled turns & 0.612 & -0.029 & 0.647 & 0.774 & -0.013 & 0.863 \\
Robot only & 0.617 & -0.024 & 0.664 & 0.785 & -0.002 & 0.875 \\
Irrelevant interleaving & 0.618 & -0.024 & 0.664 & 0.781 & -0.006 & 0.879 \\
Irrelevant suffix & 0.627 & -0.014 & 0.672 & 0.792 & 0.005 & 0.887 \\
\bottomrule
\end{tabular}
\caption{Qwen3-VL visual grounding under dialogue counterfactuals. \textbf{Full dialogue}: original unedited dialogue history from our benchmark. \textbf{mIoU}: mean IoU of the top-1 prediction; \textbf{Acc@$.5$}: accuracy at IoU $\geq 0.5$. \textbf{$\Delta\text{IoU}$}: paired mean IoU gap versus full dialogue on the same instance. Results are split into ambiguous ($n{=}232$) vs.\ non-ambiguous ($n{=}248$) instruction subsets.}
\label{tab:appendix_qwen_dialogue_interventions}
\end{table*}

\end{document}